\documentclass[runningheads]{llncs}
\usepackage[T1]{fontenc}
\usepackage{graphicx}
\usepackage{amsmath,amssymb,amsfonts}
\usepackage{algorithmic}
\usepackage{algorithm}
\usepackage{float}
\usepackage{placeins}
\usepackage{rotating}
\usepackage{multirow}
\usepackage{makecell}
\usepackage{url}          

\newcommand{\doi}[1]{\url{https://doi.org/#1}}

\begin{document}
\title{From Short Histories to Long Futures: Horizon-Aware Graph Neural Networks for Long Horizon Forecasting\thanks{Supported by NSF BIGDATA awards (IIS-1838230, IIS-2308649), NSF Leadership Class Computing awards (OAC-2139536), NSF PFI award (2423211), IBM, and Amazon.}}
\titlerunning{Horizon-Aware Graph Neural Networks for Multi-Step Forecasting}
%
\author{Zesheng Liu\inst{1}\orcidID{0009-0004-6978-7160} \and
Maryam Rahnemoonfar\inst{1,2}\orcidID{0000-0001-9358-2836}}
\authorrunning{Liu et al.}
%
\institute{Department of Computer Science and Engineering, Lehigh University, Bethlehem PA 18015, USA \and
Department of Civil and Environmental Engineering, Lehigh University, Bethlehem PA 18015, USA}
\maketitle              
\begin{abstract}
Accurate long-range prediction of geophysical systems is difficult due to strongly nonlinear dynamics, the high computational cost of full-physics simulations, and the error accumulation that arise when one-step autoregressive surrogates are rolled out over decades. Deep neural network can serve as efficient emulators, but most are trained only for next-step prediction and often drift or become unstable as the forecast horizon grows. We propose a multi-horizon graph neural network emulator that learns state-to-state transitions from a single current time to multiple future lead times within one unified model. The physical domain is represented as a graph, where nodes correspond to spatial locations with time-varying geophysical attributes and edges encode local spatial interactions. Given the current graph state, the model predicts the future evolution of key fields, ice thickness and ice velocities at all nodes, using a shared graph backbone with separate output branches for each target variable. To improve stability, the network predicts state increments relative to the current state, which are then added back to reconstruct future states. Training jointly optimizes all lead times with a unified regression objective, and inference uses a coarse-to-fine rollout that advances with larger jumps and selectively refines with shorter jumps to reduce drift and avoid redundant computation. Experiments on multi-decadal Pine Island Glacier simulations show that our approach achieves higher long-range accuracy and improved stability than both (i) an initial-state baseline that predicts each future time directly from the starting state and (ii) a standard single-step autoregressive rollout, producing a more reliable emulator for downstream climate and sea-level studies.
\keywords{Graph Neural Network \and Antarctic glaciology \and Climate Change}
\end{abstract}
%
%
%

\section{Introduction}

Accelerating, climate-driven mass loss from Greenland and Antarctica is now a leading driver of global sea-level rise, amplifying the urgency of reliable ice-sheet dynamics models \cite{Forsberg2017}. Pine Island Glacier (PIG) in West Antarctica shows the most pronounced mass loss and flow acceleration\cite{Jacobs2011,Joughin2021}, driven mainly by ocean-induced thinning near the grounding line and recurrent calving\cite{Rignot2019}. Because ice discharge is controlled by both the geometry of the ice (ice thickness) and its transport (ice velocity)\cite{Mankoff_2020_IceDischarge}, accurate prediction of ice thickness and ice velocity is central for understanding past evolution and constraining future sea-level contributions.

Classical numerical models simulate ice dynamics by solving coupled, nonlinear PDEs grounded in glaciological and thermomechanical physics. The Ice-sheet and Sea-level System Model (ISSM)\cite{Larour_2012} is a representative example that uses finite-element discretizations on unstructured meshes with adaptive refinement to resolve fast-flow regions and complex boundaries. While such numerical solvers provide high-fidelity spatiotemporal fields of different variables, they remain computationally expensive for scenario exploration: running multi-decadal transient simulations over many forcing settings (e.g., basal melt rates) requires hours to days on high-performance CPU clusters, which limits the number of scenarios that can be explored.

Machine learning emulators provide practical surrogate models: they directly learn the mapping from inputs (forcings, boundary conditions) to model states, and after a one-time training cost, they provide GPU-accelerated, batched inference with latency on the order of milliseconds to seconds, enabling large scenario sweeps, rapid what-if analyses, and near real-time decision support. A key challenge, however, is that ISSM states live on irregular, unstructured meshes rather than on regular grids. This makes standard CNN-based emulators less natural, since fixed-resolution grids can lose dynamical detail in refined fast-ice areas and waste computation in coarse slow-ice areas.

To address this challenge, Koo et al.~\cite{Koo_Rahnemoonfar_2025,Koo_Helheim} proposed a graph neural network (GNN) emulator that operates directly on the ISSM mesh: mesh vertices are treated as graph nodes and triangle connectivity induces graph edges. Their GNN forecasts the primary transient outputs of interest, the two horizontal velocity components and ice thickness, on the original unstructured mesh, and they demonstrate strong fidelity and substantial GPU speedups on transient PIG simulations across multiple mesh resolutions and basal-melt scenarios. This line of work establishes GNNs as an effective mesh-native surrogate for ISSM-style ice-sheet modeling. Despite this advantage, the first GCN emulator left sizable residuals, especially on velocity fields. In order to imporve the accuracy, Liu et al.~\cite{liu2025kangcncombiningkolmogorovarnoldnetwork} reframed emulation as residual state evolution: instead of forecasting the full physical fields at time $t$ directly from the initial states, the model learns to predict an residual update from the current state at $t-1$ to the next state at $t$. While this one-step residual formulation substantially improves short-term accuracy, deploying it autoregressively over many steps still leads to error accumulation, making long-horizon (multi-year) forecasts challenging.

In this work, we generalize the prior one-step, graph-based residual emulator to the setting of long-horizon forecasting. Specifically, given an observed prefix of a transient ISSM trajectory on an unstructured mesh, our goal is to forecast the remaining future evolution of the key physical fields, ice thickness and horizontal ice velocity, over an extended time window. The central challenge is that purely one-step models must be deployed autoregressively, so small per-step errors compound and can lead to drift over long rollouts. Our core innovation is a multi-horizon training and inference strategies that extends previous one-step residual mapping $t-1 \rightarrow t$ to a horizon-conditioned mapping $t \rightarrow t+h$. Rather than training a network that only advances the state by one time step, we train a horizon-conditioned GCN emulator that takes the current physical state on the mesh grid together with a series of discrete forecast horizon $h \in \mathcal{H}$ and predicts residual updates for velocity and thickness at time $t+h$. The horizon is encoded as an additional input feature and shared across all nodes, so a single set of GCN parameters can be used to realize multiple $h$-step emulators. 
 
During training, we predefine a discrete horizon set $\mathcal{H}$ and construct multi-horizon supervision by pairing each available state $X_t$ with its future counterparts $\{X_{t+h}\}_{h \in \mathcal{H}}$ whenever they exist in the observed simulated trajectory. The model is trained to predict the corresponding residuals $\Delta_{t,h}=X_{t+h}-X_t$ and is supervised against the numerically simulated targets at time $t+h$ across all horizons $h$ in $\mathcal{H}$, yielding a single shared-parameter emulator that learns both short- and long-lead dynamics. At inference time, we adopt a greedy descending-horizon rollout. Forecasts are generated using the largest $h \in \mathcal{H}$ first, followed by progressively smaller horizons to refine the trajectory. By prioritizing long jumps and reducing the depth of autoregression, this approach alleviates error accumulation in extended forecasts, and thereby enhances the fidelity of long-horizon emulation over extended forecast windows. Our contributions are as follows:
\begin{itemize}
    \item \textbf{Horizon-conditioned multi-horizon residual emulator.} We propose a single shared-parameter, horizon-conditioned GCN that generalizes one-step residual emulation to $t \rightarrow t+h$, forecasting ice thickness and horizontal velocity increments at multiple discrete lead times on unstructured ISSM meshes.
    
    \item \textbf{Multi-horizon supervision and joint training.} We predefine a horizon set $\mathcal{H}$ and construct training supervision by pairing each state $X_t$ with all available future states $\{X_{t+h}\}_{h \in \mathcal{H}}$, then train one shared-parameter model by minimizing a joint objective that sums residual regression losses across all horizons $h \in \mathcal{H}$.
    
    \item \textbf{Greedy descending-horizon inference.} We develop a greedy descending-horizon rollout strategy that prioritizes long jumps (largest $h$ first) and then applies progressively smaller horizons to refine the forecasted trajectory, reducing the depth of autoregression and mitigating long-horizon error accumulation.
    
    \item \textbf{Extensive evaluation on PIG ISSM transients.} We conduct extensive experiments on multi-decadal Pine Island Glacier ISSM simulations, demonstrating improved long-range accuracy and stability over single-step autoregressive rollouts and analyzing the effects of horizon-set choices and rollout strategy.
\end{itemize}


\section{Related Work}
Graph neural networks (GNNs) have been widely used for spatiotemporal modeling and digital twins due to their effectiveness in capturing both spatial and temporal dependencies in data\cite{Ngo_2023_DigitalTwinGNN}. Specifically, GNN-based emulation methods have been employed in various scientific and engineering applications, including soft-tissue mechanics~\cite{Dalton_2023_physicsinformedgnn}, electromagnetic dynamics~\cite{Noakoasteen_2024}, and greenhouse gas emissions~\cite{Fillola_2025_GATES}.

In terms of ice-sheet simulations, early deep-learning emulators~\cite{Jouvet_Cordonnier_Kim_Lüthi_Vieli_Aschwanden_2022,Jouvet_2023,Jouvet_Cordonnier_2023} largely follow the regular-grid paradigm, utilizing convolutional neural networks (CNNs) or fully-connected networks (FCNs) to predict future ice states on structured grids. However, CNNs are designed for regular grids with a fixed neighborhood pattern, while ice-sheet numerical simulations use unstructured finite-element meshes whose connectivity and resolution vary across the domain. As a result, converting these simulation outputs to a uniform grid for CNN-based emulation can miss fine details in fast-flow regions or waste computation in slow regions, motivating mesh-native alternatives.

To better match finite-element ice-sheet models, recent works have moved to graph-based emulators that operate directly on the native mesh connectivity of the ice-sheet models. Koo and Rahnemoonfar~\cite{Koo_Rahnemoonfar_2025} propose a Graph Convolutional Network (GCN) emulator for the Ice-Sheet and Sea-Level System Model (ISSM) transient simulations of Pine Island Glacier, converting triangular elements into a graph structure and predicting node-wise ice thickness and velocities. By preserving the flexible resolution of the mesh structure, their GCN emulator achieves higher fidelity than grid-based FCN and node independent MLP baselines, while enabling large GPU speedups over the original ISSM simulations for rapid sensitivity analyses~\cite{Koo_Rahnemoonfar_2025}. Beyond state emulation, graph surrogates have also been used to accelerate parameter calibration for challenging moving-boundary processes. Koo et al.\cite{Koo_Helheim} developed GNN-based emulators trained on Helheim Glacier finite-element simulations to calibrate calving parameterizations, reproducing the observed evolution of ice velocity, thickness, and ice-front migration and enabling efficient search over calving stress thresholds.

More recently, Liu et al.~\cite{liu2025kangcncombiningkolmogorovarnoldnetwork} expand the previous state emulation work by Koo and Rahnemoonfar, aiming to further improve the prediction accuracy. They propose KAN-GCN, placing a Kolmogorov-Arnold Network as a feature-wise calibrator before graph convolutions to enhance nonlinear feature conditioning while keeping the spatial inductive bias of GCNs. Additionally, they introduce a multi-head prediction design that uses separate output heads for velocity and thickness, and optimize them with two dedicated loss terms combined using tuned loss weights to balance the different scales and error sensitivities of these variables. Finally, they reframe the learning problemfrom an initial-state-to-future mapping (predicting state at time $t$ directly from the initial state) as a one-step transition model that predicts the next state at time $t$ from the previous state at time $t-1$, which better matches the transient evolution setting and supports autoregressive rollout.

While these prior works have advanced graph-based emulation for ISSM, they typically operate at a single forecasting scale by either predicting a future state from the initial or current state at a fixed lead time, or learning a one-step transition and extending it to longer horizons through repeated autoregressive rollout. This design can be fragile for long-range forecasting because small short-term errors compound over time and the model is increasingly driven by its own imperfect predictions, leading to error accumulation and distribution drift. 

In contrast, we propose a multi-horizon, horizon-conditioned GNN that learns a set of lead-time transitions within a single unified model and is trained to predict multiple future states directly from the current state. We jointly optimize all horizons during training and use a coarse-to-fine descending-horizon rollout during inference that first takes longer jumps to reach distant times and then refines intermediate states with shorter steps. This combination is designed to improve long-range stability and forecasting accuracy for transient ice-sheet emulation.


\section{Methodology}

\subsection{Graph Convolution Network}
Traditional finite-element numerical models simulates on irregular spatial discretizations where each site interacts primarily with its neighbors. Graph neural networks naturally encode this setting. They operate on arbitrary meshes, respect locality through the edges, and share parameters across nodes, making them well-suited emulators for FEM-based numerical models.

\begin{figure*}[!t]
    \centering
    \includegraphics[width=0.85\linewidth]{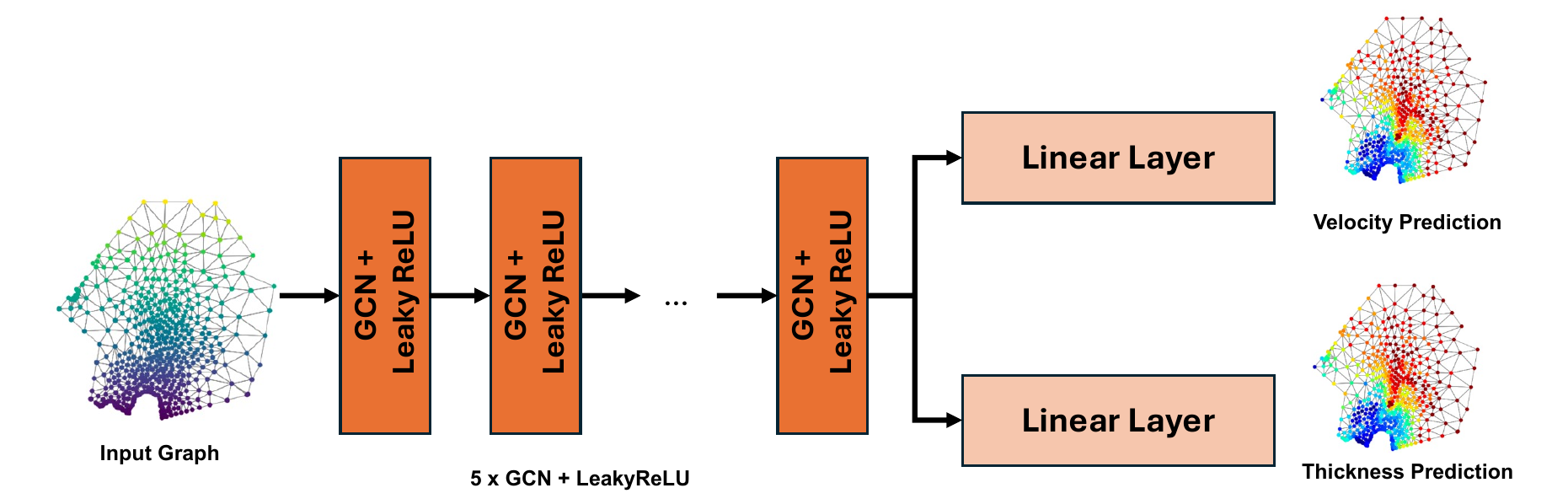}
    \caption{Network architecture of the graph neural network emulator used in this work.}
    \label{fig:network}
\end{figure*}

Figure~\ref{fig:network} shows the graph neural network emulator used in this work, which includes a stack of five Graph Convolutional Networks (GCNs)~\cite{kipf2017semi} and two separate linear prediction head. Among many graph operators, we adopt GCNs as a simple, stable building block. It can be viewed as a first-order approximation to localized spectral filters on graphs~\cite{Defferrard2016,HAMMOND2011129}, yielding an efficient neighborhood aggregation. Let \(\tilde{\mathbf{A}}=\mathbf{A}+\mathbf{I}_N\) be the adjacency with self-loops and \(\tilde{\mathbf{D}}=\mathrm{diag}(\sum_j \tilde{A}_{ij})\) the corresponding degree matrix. A GCN layer updates node features by symmetrically normalizing the aggregated messages:

\begin{equation}
\label{eq:gcn}
\mathbf{H}^{(\ell+1)}
\;=\;
\sigma\!\Big(
\tilde{\mathbf{D}}^{-\frac{1}{2}}\,
\tilde{\mathbf{A}}\,
\tilde{\mathbf{D}}^{-\frac{1}{2}}\,
\mathbf{H}^{(\ell)}\,\mathbf{W}^{(\ell)}
\Big),
\end{equation}

where \(\mathbf{H}^{(\ell)}\in\mathbb{R}^{N\times D_\ell}\) are the node embeddings at layer \(\ell\), \(\mathbf{W}^{(\ell)}\) is a trainable weight matrix, \(\sigma(\cdot)\) is a nonlinearity, and \(\mathbf{H}^{(0)}=\mathbf{X}\) (the input node features). Intuitively, the operator \(\tilde{\mathbf{D}}^{-\frac{1}{2}}\tilde{\mathbf{A}}\tilde{\mathbf{D}}^{-\frac{1}{2}}\) averages each node with its neighbors (including itself) and then applies a learnable linear transform. Deeper stacks repeat this propagate–transform step to communicate information across larger graph neighborhoods. In our spatio-temporal prediction setting, the GCN stack serves as the spatial encoder applied at each time step on the ISSM mesh graph. While Liu et al.~\cite{liu2025kangcncombiningkolmogorovarnoldnetwork} augment GCNs with Kolmogorov--Arnold Networks (KAN) to further improve representational capacity, our work focuses on the proposed multi-horizon training and greedy rollout strategy; therefore, we adopt a simpler backbone consisting of five GCN layers. Following~\cite{liu2025kangcncombiningkolmogorovarnoldnetwork}, we use two task-specific linear heads for velocity and thickness, and train the network to predict state increments (residual updates) rather than absolute states.

\subsection{Multi-Horizon Training Strategy}
\label{sec:multi_horizon}

We aim to build an emulator that forecasts ice velocity and thickness on a fixed simulation mesh represented as a graph $\mathcal{G}$. Given a trajectory $\{\mathbf{X}_1,\dots,\mathbf{X}_T\}$, a standard approach trains a one-step model ($h=1$) and deploys it autoregressively (AR): the emulator is repeatedly applied and its own predictions are fed back as inputs to reach long lead times. This AR rollout is effective in principle, but in practice it often suffers from error accumulation because small local prediction errors can compound over many steps.

Here, we propose a multi-horizon training strategy. The key idea is to replace one-step ($t\!\rightarrow\!t\!+\!1$) supervision with a horizon-conditioned, multi-step formulation that learns to forecast from an anchor time $t$ to multiple future times $t+h$. This design targets long-horizon forecasting directly: a one-step model can only reach a distant lead time by being applied repeatedly in an autoregressive rollout, which increases the number of times prediction errors can compound and gradually drives the inputs away from the ground-truth states seen during training. In contrast, multi-horizon training exposes the emulator to supervision at several lead times during optimization. By conditioning on the horizon $h$ as an explicit input and sharing parameters across all $h\in\mathcal{H}$, the same model is encouraged to produce accurate predictions both for short jumps (which capture local dynamics) and for longer jumps (which reflect accumulated change over time). As a result, long-horizon forecasting can be carried out with fewer recursive applications and with a model that has been explicitly trained to predict multi-step evolution, improving robustness when forecasting far beyond the anchor state.

We pre-define a finite set of forward horizons
$
\mathcal{H} \subset \mathbb{Z}_{>0},
$
for example $\mathcal{H}=\{1\}$ for single-step prediction or $\mathcal{H}=\{1,6,12\}$ for mixed short- and long-range jumps. For any anchor time $t$ and horizon $h\in\mathcal{H}$ such that $1\le t\le T-h$, we predict the $h$-step increment
$
\Delta\hat{\mathbf{X}}_{t\rightarrow t+h}
=
G_\theta\big(\mathbf{X}_t,\mathbf{F}_t,\psi(h),\mathcal{G}\big),
$
and recover the absolute target state on the prognostic channels via
$
\hat{\mathbf{X}}_{t+h}
=
\mathbf{X}_t+\Delta\hat{\mathbf{X}}_{t\rightarrow t+h},
$
following a residual formulation~\cite{liu2025kangcncombiningkolmogorovarnoldnetwork}.
Here $\mathbf{X}_t$ is the prognostic state at the anchor time $t$ (restricted to the channels we evolve, e.g., velocity and thickness), $\mathbf{F}_t$ collects forcing/context features available at time $t$, and $\psi(h)\in\mathbb{R}$ is a normalized horizon encoding. In our implementation, we use the scalar encoding
$
\psi(h)=h_{\mathrm{norm}}=\frac{h}{H_{\max}}
$
which is broadcast and concatenated with the per-node input features and 
$
H_{\max}=\max_{h \in \mathcal{H}} h
$.

As shown in Figure~\ref{fig:network}, a shared stack of graph convolutional layers on $\mathcal{G}$ maps the combined input $(\mathbf{X}_t,\mathbf{F}_t,\psi(h))$ to node embeddings, and two linear heads transform these embeddings into velocity and thickness deltas. The same parameter set $G_\theta$ is used for all horizons in $\mathcal{H}$; no separate network is instantiated for each $h$. Training pairs are constructed by enumerating all valid anchor--horizon combinations
$
\mathcal{D}=\{(t,h): h\in\mathcal{H},\ 1\le t\le T-h\},
$
with supervision given by the ground-truth multi-step difference
$
\Delta\mathbf{X}_{t \rightarrow t+h}
=
\mathbf{X}_{t+h} - \mathbf{X}_t
$
restricted to the velocity and thickness channels. During training, we minimize a mean-squared error on the prognostic channels, with separate terms for velocity and thickness:
\[
\mathcal{L}
=
\frac{1}{|\mathcal{D}|}
\sum_{(t,h)\in\mathcal{D}}
\left(
\lambda_v \big\|\hat{\mathbf{V}}_{t+h}-\mathbf{V}_{t+h}\big\|_2^2
+
\lambda_H \big\|\hat{\mathbf{H}}_{t+h}-\mathbf{H}_{t+h}\big\|_2^2
\right),
\]
where $\mathbf{V}_{t+h}$ and $\mathbf{H}_{t+h}$ denote the ground-truth velocity and thickness at time $t+h$ (stacked over all nodes), and $\hat{\mathbf{V}}_{t+h}$ and $\hat{\mathbf{H}}_{t+h}$ are the corresponding predictions. The weights $(\lambda_v,\lambda_H)$ balance the different magnitudes of velocity and thickness.

\subsection{Greedy Descending-Horizon Inference Strategy}
\label{sec:greedy_rollout}

We focus on long-horizon forecasting on a held-out simulation trajectory of length $T$, denoted by $\{\mathbf{X}_1,\ldots,\mathbf{X}_T\}$, defined on the same graph $\mathcal{G}$. Based on the multi-horizon training framework, we design a greedy descending-horizon rollout strategy for long-range forecasting, which combines short and long jumps to efficiently cover an extended forecast window. Let $t_0$ denote the last available (observed) time index. We assume the initial prefix $\{\mathbf{X}_t\}_{t\le t_0}$ is known exactly and is used as the starting condition, while the goal is to forecast the future window $t=t_0+1,\dots,t_1$. We also assume the forcing/context inputs $\mathbf{F}_t$ are available for all $t\in[1,t_1]$ (they are exogenous inputs provided during the simulation).

Given a trained multi-horizon network $G_\theta$ and a forward horizon set $\mathcal{H}\subset\mathbb{Z}_{>0}$ (with $1\in\mathcal{H}$), we generate forecasts in the target window using a greedy descending-horizon procedure. We maintain predicted states $\{\hat{\mathbf{X}}_t\}$ and a Boolean indicator $\textit{known}(t)$ denoting whether $\hat{\mathbf{X}}_t$ is currently available. We initialize $\hat{\mathbf{X}}_t\leftarrow \mathbf{X}_t$ and $\textit{known}(t)\leftarrow \texttt{true}$ for all $t\le t_0$, and set $\textit{known}(t)\leftarrow \texttt{false}$ for $t_0<t\le t_1$.

The rollout iterates horizons from largest to smallest. For each horizon $h$, we scan forward in time and apply the emulator whenever an input time $t_{\mathrm{in}}$ is known and its target $t_{\mathrm{out}}=t_{\mathrm{in}}+h$ lies within the forecast window and is still unknown. Each predicted target is filled once and never overwritten. Because $1\in\mathcal{H}$, any remaining gaps can always be filled at the end by unit-step transitions, while larger horizons reduce the number of recursive model applications needed to cover the window. 

After the rollout, we evaluate forecast accuracy over the window $t=t_0+1,\dots,t_1$ by comparing $\hat{\mathbf{X}}_t$ with the ground truth $\mathbf{X}_t$. For each variable (the two horizontal velocity components and thickness), we first compute a per-time-step RMSE by pooling squared errors over all nodes at time $t$. We then report the window-averaged RMSE as the root mean squared error pooled over all nodes and all time steps in $t_0+1,\dots,t_1$, and convert the result to physical units.

Combining multi-horizon training (Section~\ref{sec:multi_horizon}) with this greedy descending-horizon rollout improves long-horizon forecasting in two ways. First, larger horizons reduce the number of recursive model applications required to reach distant times, lowering opportunities for compounding errors. Second, because the same network is trained to predict multi-step changes for multiple $h\in\mathcal{H}$, it is explicitly optimized to make accurate long jumps from anchor states, improving robustness when forecasting far beyond the observed prefix.


\section{Experiment Setup}

\subsection{Data Preparation}

In order to generate datasets for our proposed multi-horizon graph-based emulator, we implement transient simulations with the Ice-sheet and Sea-level System Model, a finite-element thermomechanical ice-sheet model that solves the Shelfy-Stream Approximation (SSA) on unstructured meshes with adaptive mesh refinement~\cite{Larour_2012,Seroussi_2014,Morland_1987}. For Pine Island Glacier (PIG), Antarctica, we simulate the 20-year evolution of ice thickness and velocities by adapting previously published ISSM sensitivity experiments~\cite{Larour_2012,Seroussi_2014}. Given that PIG has undergone some of the most rapid acceleration and mass loss in Antarctica and currently contributes more than 20\% of the continent's sea-level rise signal~\cite{Jacobs2011,Joughin2021,Rignot2019}, this glacier provides a ideal testbed for evaluating whether our learned emulator can reduce uncertainty in projections of ice-sheet mass loss.   

To study how emulator performance depends on spatial resolution, we follow the ISSM mesh design in recent GCN-based emulators for PIG and construct three sets of adaptive triangular meshes, initialized with nominal element sizes of 2km, 5km, and 10km~\cite{Koo_Rahnemoonfar_2025}. In each case, an anisotropic mesh generator refines the grid near dynamically important regions such as the central fast-flowing ice stream and grounding line, and coarsens it in slower, nearly stagnant ice~\cite{Larour_2012}. The resulting meshes contain on the order of $10^3$–$10^4$ elements (e.g., roughly 6000, 2800, and 1800 nodes for the 2km, 5km, and 10km configurations, respectively), thereby preserving the computational efficiency and spatial adaptivity of the underlying finite-element model~\cite{Koo_Rahnemoonfar_2025}.

Motivated by evidence that basal melting beneath the ice shelf is the dominant driver of PIG mass loss~\cite{Jacobs2011,Joughin2021}, we prescribe spatially distributed annual basal melt rates that are constant in time but varied across experiments. Specifically, we explore 36 scenarios with uniform mean melt rates from $0$ to $70\,\mathrm{m\,a^{-1}}$ in increments of $2\,\mathrm{m\,a^{-1}}$ and integrate each experiment forward for 20 years with a monthly time step, yielding 240 model states per scenario.\cite{Seroussi_2014,Koo_Rahnemoonfar_2025}. In total, this yields \(108\) simulations (three mesh sizes, each paired with \(36\) melt rates) over 20 years. All the ISSM simulations are initialized using the same observational and climatological datasets empolyed in the recent work by Koo et al.~\cite{Koo_Rahnemoonfar_2025}. We convert each ISSM triangular mesh into a graph by taking mesh vertices as nodes and mesh adjacencies as edges. Each node is attributed with static features such as melting rate and surface mass balance, as well as time-varying features include surface elevation, base elevation, ice floating ratio, combined magnitudes of ice velocities, and normalized input time state t and an encoded horizon h.

\subsection{Training Details}
For Pine Island Glacier, we divide the whole transient simulation trajectories into training, validation, and testing dataset based on their melting rate values. Specifically, trajectories with melting rate of of 0, 20, 40 and 60 $m\cdot a^{-1}$ are used for validation, trajectories with melting rate of 10, 30, 50, and 70 $m\cdot a^{-1}$ are used for testing, and the remaining are used for training. During the training phase, for each h in the horizon sets, we generated all the pairs of (t, h) from the training trajectories where t + h does not exceed the total simulation time. The same procedure is applied to the validation and testing datasets to generate (t, h) pairs for model evaluation.

All the networks are trained using the same GPU instance\cite{Jetstream2_1,Jetstream2_2}, which is composed of AMD EPYC 7713 64-Core Processor, 480 GB of system RAM, and 4 NVIDIA A100 GPUs with 40GB of VRAM. Adam\cite{kingma2017adammethodstochasticoptimization} is used as the optimizer. We use the Adam optimizer with an initial learning rate of 0.001, a weight decay factor of 0.0001, and a cosine anneal learning rate scheduler\cite{loshchilov2017sgdrstochasticgradientdescent}. To ensure the fully convergence, we train all the emulators for 500 epochs.  


\section{Results Analysis}
\subsection{Overall Performance}

\begin{figure*}[!b]
    \centering
    \includegraphics[width=0.9\linewidth]{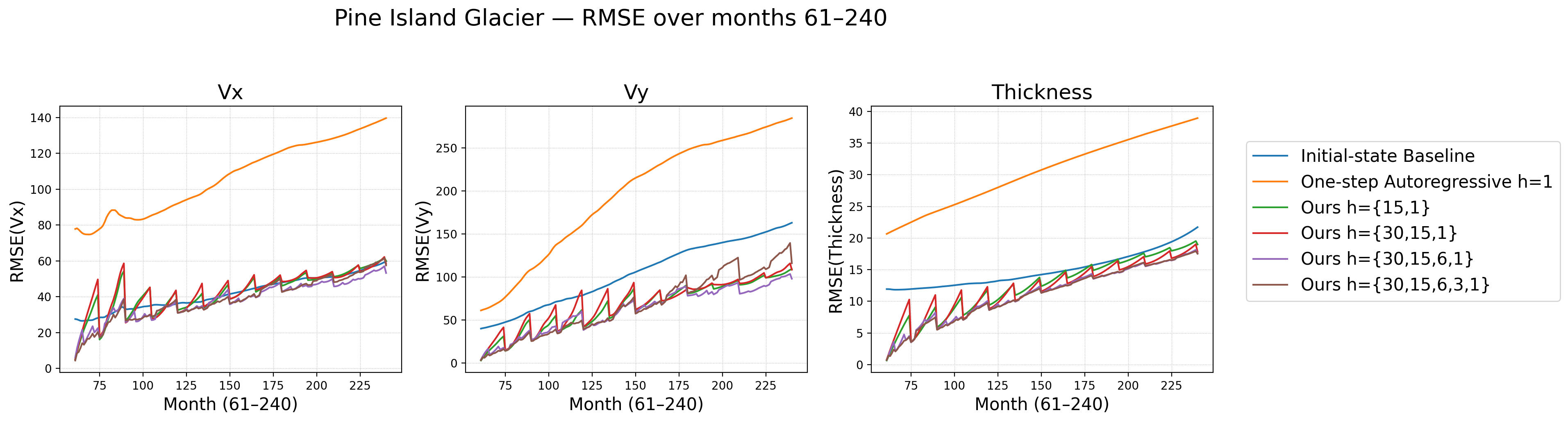}
    \caption{Per-month pooled RMSE on Pine Island Glacier forecasts (months 61--240) for $V_x$, $V_y$, and ice thickness.
    Each curve corresponds to a different method: the initial-state baseline~\cite{Koo_Rahnemoonfar_2025}, the one-step autoregressive model~\cite{liu2025kangcncombiningkolmogorovarnoldnetwork}, and our proposed multi-horizon training and greedy descending-horizon inference with different horizon sets. The figure illustrates how multi-horizon training substantially reduces both the level and growth rate of error over long roll-outs compared to the single-horizon baseline.}

    \label{fig:RMSE-Comparison}
\end{figure*}

We compare our proposed multi-horizon training and greedy descending-horizon inference strategy and against existing Pine Island Glacier emulators, including an initial-state baseline that directly predicts each future month from the inital state at month 1 (Koo and Rahnemoonfar~\cite{Koo_Rahnemoonfar_2025}), and a standard one-step emulator trained with $t \rightarrow t+1 $ supervision and deployed via autoregressive rollout for long-horizon forecast (Liu et al.~\cite{liu2025kangcncombiningkolmogorovarnoldnetwork}). To isolate the impact of the training/inference strategy from architectural differences, we use a common backbone across methods: five GCN layers as the feature extractor, followed by linear output heads for ice thickness and horizontal velocity. In particular, while Liu et al.~\cite{liu2025kangcncombiningkolmogorovarnoldnetwork} propose a KAN-GCN variant, we keep the backbone fixed here so that performance differences primarily reflect the supervision horizons and rollout procedure rather than additional architectural components.



We evaluate all methods under the same long-horizon forecasting protocol on held-out 240-month ISSM transient simulation trajectories. For each test trajectory, we assume the ground-truth states for months 1--60 are observed and our goal is to provide accurate forecasts for month 61-240. Each monthly state consists of node-wise fields on the ISSM mesh with three predicted channels $c\in\{V_x,V_y,H\}$, and different test trajectories may have different mesh resolutions (numbers of nodes). For each forecast month $t$ and channel $c$, we compute an summarized average RMSE by aggregating squared errors over all mesh nodes and all test trajectories and normalizing by the total number of nodes across the test set (i.e., a node-count-weighted average across trajectories), followed by a square root. This produces a per-month RMSE curve for each channel. All reported errors are computed exclusively on the forecast window $t=61,\ldots,240$.

As shown in Table~\ref{tab:results} and Figure~\ref{fig:RMSE-Comparison}, the standard one-step autoregressive emulator performs poorly for long-horizon forecasting. Its average RMSE error is substantially higher than all other methods across $V_x$, $V_y$, and $H$. This behavior is consistent with error accumulation under free-running rollouts, where small one-step biases quickly compound as the predicted states drift away from the training distribution. In contrast, the initial-state baseline~\cite{Koo_Rahnemoonfar_2025} predicts each future month $t$ directly from the same initial condition at month 1 (i.e., without conditioning on intermediate predicted states). It is competitive on $V_x$ but remains relatively weak on $V_y$ and thickness, suggesting that a fixed start-to-future mapping can under-represent transient dynamics and state-dependent evolution that are important for accurately tracking the changing velocity field.

Our proposed multi-horizon training and greedy descending-horizon inference strategy substantially outperforms both baselines across all channels once a properly chosen mid-range horizon is included. By providing explicit supervision for multi-step transitions, the emulator becomes less prone to compounding errors in free-running rollout, resulting in markedly lower RMSE over the full 15-year forecast window. Specifically, with only two horizons, $\mathcal{H}=\{1,15\}$ already yields a strong improvement over the one-step autoregressive baseline on long-horizon forecasting, substantially reducing errors on $V_y$ and $H$ (from $207.05\rightarrow 70.85$ and $30.91\rightarrow 13.22$ respectively), while keeping $V_x$ comparable to the initial-state baseline. The results demonstrate that multi-horizon supervision is a powerful tool for enhancing long-horizon forecasting performance in autoregressive emulators. Figure~\ref{fig:RMSE-Comparison} further shows that our multi-horizon variants not only lower the overall RMSE level but also suppress the growth of error over time, indicating improved rollout stability compared to the one-step autoregressive baseline.

We also find that adding more horizons provides further but diminishing improvements, and the gains are not always uniform across channels. Moving from $\{1,15\}$ to $\{1,15,30\}$ slightly improves thickness accuracy with comparable velocity errors. The best overall velocity performance is achieved by $\mathcal{H}=\{1,6,15,30\}$, which attains the lowest average RMSE error for $V_x$ and $V_y$ (39.57 and 66.50), and also improves thickness to 11.99. Extending the horizon set to $\{1,3,6,15,30\}$ yields the best thickness (11.91) but slightly worsens velocity compared to $\{1,6,15,30\}$, illustrating a practical trade-off: adding additional short horizons can further regularize thickness predictions, yet may redistribute optimization capacity in a way that does not consistently benefit both velocity components.

These accuracy trends come with an expected cost increase. As the horizon set grows, the number of supervised pairs $(t,h)$ increases, and we observe a monotonic rise in training time. Taken together, the results suggest that most of the improvement is obtained by introducing a single well-chosen mid-range horizon ($h_2=15$), while additional horizons offer incremental refinements that should be selected based on the target metric (velocity vs.\ thickness) and the available training budget.

\begin{table}[!b]
\centering
\small
\setlength{\tabcolsep}{4pt}
\renewcommand{\arraystretch}{1.15}
\caption{Average RMSE on the long-horizon forecast window (months 61--240). Months 1--60 are provided as observed context (warm start), and each model is rolled out in a free-running autoregressive forecast; RMSE is aggregated over all nodes and test trajectories and then averaged over $t=61,\ldots,240$ (lower is better).}
\label{tab:results}
\scalebox{0.7}{
\begin{tabular}{|p{3.4cm}|c|c|c|c|}
\hline
\textbf{Method} & \textbf{Train Time} (s) & \textbf{RMSE($V_x$)} & \textbf{RMSE($V_y$)} & \textbf{RMSE($H$)} \\ \hline
\makecell[l]{Initial-state\\Baseline~\cite{Koo_Rahnemoonfar_2025}} & / & 43.04 & 110.08 & 15.16 \\ \hline
\makecell[l]{One-step\\Autoregressive\\ ($h{=}1$)~\cite{liu2025kangcncombiningkolmogorovarnoldnetwork}} & 2388.17  & 108.77 & 207.05 & 30.91 \\ \hline
\makecell[l]{Ours ($h{=}1,15$)} & 4742.19 & 43.83  & 70.85 & 13.22 \\ \hline
\makecell[l]{Ours ($h{=}1,15,30$)} & 7074.82 & 44.45 & 73.86 & 12.76 \\ \hline
\makecell[l]{Ours ($h{=}1,6,15,30$)} & 9279.70 &\textbf{39.57} &\textbf{66.50} &11.99 \\ \hline
\makecell[l]{Ours ($h{=}1,3,6,15,30$)} &11491.40 &40.32 &75.92 &\textbf{11.91} \\ \hline
\end{tabular}
}
\end{table}

\subsection{Discussion on the Choice of Horizon Sets}

In order to guarantee the local refinement of all future states during the greedy descending-horizon rollout, the forward horizon set $\mathcal{H}$ must include $1$ as a member. Beyond this requirement, the choice of additional horizons in $\mathcal{H}$ affects both training complexity and stability, together with long-horizon forecasting performance. Including more horizons increases the number of supervision pairs $(t,h)$ extracted from each training trajectory, which can improve the model's ability to learn multi-step dynamics and therefore have better accuracy for long-horizon forecasting. However, larger horizon sets also increase the training complexity, as each training batch must now accommodate multiple horizons, leading to longer training times and higher computational costs. In certain cases, adding very large $h$ in the horizon sets(i.e. $H={1, 60}$) may also introduce optimization challenges, as the model must learn to accurately predict over a wider range of time scales, which can complicate the learning dynamics. 

\begin{table}[!b]
\centering
\caption{Ablation study on the choice of the second horizon in 2-horizon training ($\mathcal{H}=\{1,h_2\}$). We report the average RMSE over the free-running rollout window $t=61$--$240$ months (lower is better).}
\label{table:ablation-horizons}
\setlength{\tabcolsep}{6pt}
\renewcommand{\arraystretch}{1.15}
\scalebox{0.73}{
\begin{tabular}{|c|c|c|c|}
\hline
\textbf{$h_2$} & \textbf{RMSE($V_x$)} & \textbf{RMSE($V_y$)} & \textbf{RMSE($H$)} \\
\hline
2  & 64.36 & 109.18 & 17.34 \\ \hline
3  & 63.46 & 111.86 & 16.28 \\ \hline
4  & 53.51 & 178.20 & 15.32 \\ \hline
6  & 49.82 & 102.14 & 14.53 \\ \hline
8  & 54.91 & 142.16 & 14.69 \\ \hline
9  & 47.08 & 97.03 & 13.76 \\ \hline
12 & 45.89 & 76.58 & 13.72 \\ \hline
15 & \textbf{43.83} & \textbf{70.85} & \textbf{13.22} \\ \hline
18 & 47.13 & 80.27 & 13.31 \\ \hline
24 & 53.69 & 79.24 & 13.73 \\ \hline
36 & 56.03 & 85.39 & 14.80 \\ \hline
\end{tabular}
}
\end{table}

Table~\ref{table:ablation-horizons} presents an ablation study on the choice of second member in the horizon sets for 2-horizon training. We fix the first horizon as $1$ and vary the second horizon from $\{2,3,4,6,8,9,12,15,18,24, 36\}$. For each configuration, we report the average RMSE over the 61--240 months rollout. Several trends emerge from the experiments. First, adding a moderately long horizon yields a substantial improvement over short horizons: configurations with $h_2\in\{12,15,18\}$ consistently outperform $h_2\le 6$ on all three channels. In particular, $h_2=15$ achieves the best overall performance among the candidates, reducing RMSE to $43.83$ ($V_x$), $70.85$ ($V_y$), and $13.22$ ($H$). This indicates that most of the benefit of multi-horizon supervision comes from introducing a second horizon that is sufficiently long to expose the model to meaningful multi-month dynamics, while still remaining within a stable and learnable range.

Second, extremely short or poorly matched horizons can be counterproductive for specific channels. For example, although $h_2=4$ improves $V_x$, it substantially degrades $V_y$, suggesting that supervision at a single intermediate but short lead time may bias the optimization toward certain variables without improving overall rollout stability. Likewise, very large horizons can increase optimization difficulty and sensitivity to modeling errors. While $h_2=36$ does not significantly diverge in this 2-horizon setting, it is clearly worse than the best mid-range choices, supporting the intuition that horizons that are too long may make training harder without commensurate gains. Therefore, selecting an appropriate mid-range second horizon is crucial for balancing training complexity and long-horizon forecasting performance.

\section{Conclusion}

In this work, we extened existing residual, graph-based emulators of the Ice-sheet and Sea-level System Model to explicitly support multiple forecast lead times for Pine Island Glacier. Building on single-step residual GCN emulators, we introduce a horizon-conditioned graph network that takes the physical state on the PIG mesh at time $t$ together with a discrete lead time $h$ and predicts residual updates for velocity and thickness at time $t+h$. We encode $h$ as a simple normalized scalar and concatenate it to the node features, enabling one shared GCN backbone to operate across a set of lead times. This yields a unified training formulation in which, for each transient ISSM trajectory, we construct all valid supervision pairs $(t,h)$ from a predefined horizon set $\mathcal{H}$ and train the model to match multi-step residual targets produced by the numerical simulator. 

Crucially, we propose two complementary strategies for long-horizon forecasting. First, we propose a multi-horizon training strategy that replaces a purely one-step objective with joint supervision across multiple lead times, so the model learns to make accurate direct predictions at longer horizons instead of relying exclusively on repeated composing one-step predictions. This reduces error compounding and mitigates distribution drift during long autoregressive roll-outs. Second, we introduce a greedy descending-horizon inference stratgey, which performs a long-window rollout by combining long and short jumps: larger horizons provide efficient progress and establish coarse future states, while shorter horizons refine the trajectory and act as local corrections, without overwriting already-produced predictions. Together, these design choices retain strong short-lead accuracy while improving stability and robustness in long-horizon forecasts.

Experiments on transient Pine Island Glacier simulations demonstrate that the proposed multi-horizon residual emulator can remain stable over extended forecast windows and offers a practical trade-off between robustness (short corrective jumps) and computational efficiency (long jumps). In particular, under the long-horizon evaluation protocol where months 1--60 are treated as observed context and errors are reported only on the forecast window $t=61,\ldots,240$, our proposed multi-horizon training and greedy descending-horizon inference strategies achieve the lowest error and best stability compared to single-step residual emulators and initial-state baselines, highlighting their effectiveness for long-range forecasting by combining direct long-step predictions with short-step refinement.

\subsubsection{Acknowledgements} This work is supported by NSF BIGDATA awards (IIS-1838230, IIS-2308649), NSF Leadership Class Computing awards (OAC-2139536), NSF PFI awards (2423211).

%
%
%
\bibliographystyle{splncs04}
\bibliography{ref}

@article{Koo_Rahnemoonfar_2025, title={Graph convolutional network as a fast statistical emulator for numerical ice sheet modeling}, volume={71}, DOI={10.1017/jog.2024.93}, journal={Journal of Glaciology}, author={Koo, Younghyun and Rahnemoonfar, Maryam}, year={2025}, pages={e15}}

@inproceedings{Jetstream2_1,
author = {Hancock, David Y. and Fischer, Jeremy and Lowe, John Michael and Snapp-Childs, Winona and Pierce, Marlon and Marru, Suresh and Coulter, J. Eric and Vaughn, Matthew and Beck, Brian and Merchant, Nirav and Skidmore, Edwin and Jacobs, Gwen},
title = {Jetstream2: Accelerating cloud computing via Jetstream},
year = {2021},
isbn = {9781450382922},
publisher = {Association for Computing Machinery},
address = {New York, NY, USA},
url = {https://doi.org/10.1145/3437359.3465565},
doi = {10.1145/3437359.3465565},
booktitle = {Practice and Experience in Advanced Research Computing 2021: Evolution Across All Dimensions},
articleno = {11},
numpages = {8},
location = {Boston, MA, USA},
series = {PEARC '21}
}

@inproceedings{Jetstream2_2,
author = {Boerner, Timothy J. and Deems, Stephen and Furlani, Thomas R. and Knuth, Shelley L. and Towns, John},
title = {ACCESS: Advancing Innovation: NSF’s Advanced Cyberinfrastructure Coordination Ecosystem: Services \& Support},
year = {2023},
isbn = {9781450399852},
publisher = {Association for Computing Machinery},
address = {New York, NY, USA},
url = {https://doi.org/10.1145/3569951.3597559},
doi = {10.1145/3569951.3597559},
booktitle = {Practice and Experience in Advanced Research Computing 2023: Computing for the Common Good},
pages = {173–176},
numpages = {4},
location = {Portland, OR, USA},
series = {PEARC '23}
}

@misc{kingma2017adammethodstochasticoptimization,
      title={Adam: A Method for Stochastic Optimization}, 
      author={Diederik P. Kingma and Jimmy Ba},
      year={2017},
      eprint={1412.6980},
      archivePrefix={arXiv},
      primaryClass={cs.LG},
      url={https://arxiv.org/abs/1412.6980}, 
}

@misc{loshchilov2017sgdrstochasticgradientdescent,
      title={SGDR: Stochastic Gradient Descent with Warm Restarts}, 
      author={Ilya Loshchilov and Frank Hutter},
      year={2017},
      eprint={1608.03983},
      archivePrefix={arXiv},
      primaryClass={cs.LG},
      url={https://arxiv.org/abs/1608.03983}, 
}

@Article{Jacobs2011,
author={Jacobs, Stanley S.
and Jenkins, Adrian
and Giulivi, Claudia F.
and Dutrieux, Pierre},
title={Stronger ocean circulation and increased melting under Pine Island Glacier ice shelf},
journal={Nature Geoscience},
year={2011},
month={Aug},
day={01},
volume={4},
number={8},
pages={519-523},
issn={1752-0908},
doi={10.1038/ngeo1188},
url={https://doi.org/10.1038/ngeo1188}
}

@article{Joughin2021,
author = {Ian Joughin  and Daniel Shapero  and Pierre Dutrieux  and Ben Smith },
title = {Ocean-induced melt volume directly paces ice loss from Pine Island Glacier},
journal = {Science Advances},
volume = {7},
number = {43},
pages = {eabi5738},
year = {2021},
doi = {10.1126/sciadv.abi5738},
URL = {https://www.science.org/doi/abs/10.1126/sciadv.abi5738},
eprint = {https://www.science.org/doi/pdf/10.1126/sciadv.abi5738}}

@inproceedings{kipf2017semi,
  title={Semi-Supervised Classification with Graph Convolutional Networks},
  author={Kipf, Thomas N. and Welling, Max},
  booktitle={International Conference on Learning Representations (ICLR)},
  year={2017}
}

@article{HAMMOND2011129,
title = {Wavelets on graphs via spectral graph theory},
journal = {Applied and Computational Harmonic Analysis},
volume = {30},
number = {2},
pages = {129-150},
year = {2011},
issn = {1063-5203},
doi = {https://doi.org/10.1016/j.acha.2010.04.005},
url = {https://www.sciencedirect.com/science/article/pii/S1063520310000552},
author = {David K. Hammond and Pierre Vandergheynst and Rémi Gribonval}
}

@inproceedings{Defferrard2016,
author = {Defferrard, Micha\"{e}l and Bresson, Xavier and Vandergheynst, Pierre},
title = {Convolutional neural networks on graphs with fast localized spectral filtering},
year = {2016},
isbn = {9781510838819},
publisher = {Curran Associates Inc.},
address = {Red Hook, NY, USA},
booktitle = {Proceedings of the 30th International Conference on Neural Information Processing Systems},
pages = {3844–3852},
numpages = {9},
location = {Barcelona, Spain},
series = {NIPS'16}
}

@Article{Seroussi_2014,
AUTHOR = {Seroussi, H. and Morlighem, M. and Rignot, E. and Mouginot, J. and Larour, E. and Schodlok, M. and Khazendar, A.},
TITLE = {Sensitivity of the dynamics of Pine Island Glacier, West Antarctica, to climate forcing for the next 50 years},
JOURNAL = {The Cryosphere},
VOLUME = {8},
YEAR = {2014},
NUMBER = {5},
PAGES = {1699--1710},
URL = {https://tc.copernicus.org/articles/8/1699/2014/},
DOI = {10.5194/tc-8-1699-2014}
}

@article{Larour_2012,
author = {Larour, E. and Schiermeier, J. and Rignot, E. and Seroussi, H. and Morlighem, M. and Paden, J.},
title = {Sensitivity Analysis of Pine Island Glacier ice flow using ISSM and DAKOTA},
journal = {Journal of Geophysical Research: Earth Surface},
volume = {117},
number = {F2},
pages = {},
doi = {https://doi.org/10.1029/2011JF002146},
url = {https://agupubs.onlinelibrary.wiley.com/doi/abs/10.1029/2011JF002146},
eprint = {https://agupubs.onlinelibrary.wiley.com/doi/pdf/10.1029/2011JF002146},
year = {2012}
}

@InProceedings{Morland_1987,
author="Morland, L. W.",
editor="Van der Veen, C. J.
and Oerlemans, J.",
title="Unconfined Ice-Shelf Flow",
booktitle="Dynamics of the West Antarctic Ice Sheet",
year="1987",
publisher="Springer Netherlands",
address="Dordrecht",
pages="99--116",
isbn="978-94-009-3745-1"
}

@Inbook{Forsberg2017,
author="Forsberg, Rene
and Sorensen, Louise
and Simonsen, Sebastian",
title="Greenland and Antarctica Ice Sheet Mass Changes and Effects on Global Sea Level",
bookTitle="Integrative Study of the Mean Sea Level and Its Components",
year="2017",
publisher="Springer International Publishing",
address="Cham",
pages="91--106",
isbn="978-3-319-56490-6",
doi="10.1007/978-3-319-56490-6_5",
url="https://doi.org/10.1007/978-3-319-56490-6_5"
}

@Article{Koo_Helheim,
AUTHOR = {Koo, Y. and Cheng, G. and Morlighem, M. and Rahnemoonfar, M.},
TITLE = {Calibrating calving parameterizations using graph neural network emulators: application to Helheim Glacier, East Greenland},
JOURNAL = {The Cryosphere},
VOLUME = {19},
YEAR = {2025},
NUMBER = {7},
PAGES = {2583--2599},
URL = {https://tc.copernicus.org/articles/19/2583/2025/},
DOI = {10.5194/tc-19-2583-2025}
}

@article{Rignot2019,
author = {Eric Rignot  and Jérémie Mouginot  and Bernd Scheuchl  and Michiel van den Broeke  and Melchior J. van Wessem  and Mathieu Morlighem },
title = {Four decades of Antarctic Ice Sheet mass balance from 1979–2017},
journal = {Proceedings of the National Academy of Sciences},
volume = {116},
number = {4},
pages = {1095-1103},
year = {2019},
doi = {10.1073/pnas.1812883116},
URL = {https://www.pnas.org/doi/abs/10.1073/pnas.1812883116},
eprint = {https://www.pnas.org/doi/pdf/10.1073/pnas.1812883116}}

@misc{liu2025kangcncombiningkolmogorovarnoldnetwork,
      title={KAN-GCN: Combining Kolmogorov-Arnold Network with Graph Convolution Network for an Accurate Ice Sheet Emulator}, 
      author={Zesheng Liu and YoungHyun Koo and Maryam Rahnemoonfar},
      year={2025},
      eprint={2510.24926},
      archivePrefix={arXiv},
      primaryClass={cs.LG},
      url={https://arxiv.org/abs/2510.24926}, 
}

@Article{Mankoff_2020_IceDischarge,
AUTHOR = {Mankoff, K. D. and Solgaard, A. and Colgan, W. and Ahlstr{\o}m, A. P. and Khan, S. A. and Fausto, R. S.},
TITLE = {Greenland Ice Sheet solid ice discharge from 1986 through March 2020},
JOURNAL = {Earth System Science Data},
VOLUME = {12},
YEAR = {2020},
NUMBER = {2},
PAGES = {1367--1383},
URL = {https://essd.copernicus.org/articles/12/1367/2020/},
DOI = {10.5194/essd-12-1367-2020}
}

@Article{Ngo_2023_DigitalTwinGNN,
AUTHOR = {Ngo, Duc-Thinh and Aouedi, Ons and Piamrat, Kandaraj and Hassan, Thomas and Raipin-Parvédy, Philippe},
TITLE = {Empowering Digital Twin for Future Networks with Graph Neural Networks: Overview, Enabling Technologies, Challenges, and Opportunities},
JOURNAL = {Future Internet},
VOLUME = {15},
YEAR = {2023},
NUMBER = {12},
ARTICLE-NUMBER = {377},
URL = {https://www.mdpi.com/1999-5903/15/12/377},
ISSN = {1999-5903},
DOI = {10.3390/fi15120377}
}

@article{Dalton_2023_physicsinformedgnn,
title = {Physics-informed graph neural network emulation of soft-tissue mechanics},
journal = {Computer Methods in Applied Mechanics and Engineering},
volume = {417},
pages = {116351},
year = {2023},
issn = {0045-7825},
doi = {https://doi.org/10.1016/j.cma.2023.116351},
url = {https://www.sciencedirect.com/science/article/pii/S0045782523004759},
author = {David Dalton and Dirk Husmeier and Hao Gao}
}

@article{Noakoasteen_2024,
author = {Noakoasteen, O. and Christodoulou, C. and Peng, Z. and Goudos, S. K.},
title = {Physics-informed surrogates for electromagnetic dynamics using Transformers and graph neural networks},
journal = {IET Microwaves, Antennas \& Propagation},
volume = {18},
number = {7},
pages = {505-515},
doi = {https://doi.org/10.1049/mia2.12463},
url = {https://ietresearch.onlinelibrary.wiley.com/doi/abs/10.1049/mia2.12463},
eprint = {https://ietresearch.onlinelibrary.wiley.com/doi/pdf/10.1049/mia2.12463},
year = {2024}
}

@Article{Fillola_2025_GATES,
AUTHOR = {Fillola, E. and Santos-Rodriguez, R. and Tunnicliffe, R. and Clark, J. and Keshtmand, N. and Ganesan, A. and Rigby, M.},
TITLE = {Enabling Fast Greenhouse Gas Emissions Inference from Satellites with GATES: a Graph-Neural-Network Atmospheric Transport Emulation System},
JOURNAL = {EGUsphere},
VOLUME = {2025},
YEAR = {2025},
PAGES = {1--27},
URL = {https://egusphere.copernicus.org/preprints/2025/egusphere-2025-2392/},
DOI = {10.5194/egusphere-2025-2392}
}

@article{Jouvet_Cordonnier_Kim_Lüthi_Vieli_Aschwanden_2022, title={Deep learning speeds up ice flow modelling by several orders of magnitude}, volume={68}, DOI={10.1017/jog.2021.120}, number={270}, journal={Journal of Glaciology}, author={Jouvet, Guillaume and Cordonnier, Guillaume and Kim, Byungsoo and Lüthi, Martin and Vieli, Andreas and Aschwanden, Andy}, year={2022}, pages={651–664}}

@article{Jouvet_2023, title={Inversion of a Stokes glacier flow model emulated by deep learning}, volume={69}, DOI={10.1017/jog.2022.41}, number={273}, journal={Journal of Glaciology}, author={Jouvet, Guillaume}, year={2023}, pages={13–26}}

@article{Jouvet_Cordonnier_2023, title={Ice-flow model emulator based on physics-informed deep learning}, volume={69}, DOI={10.1017/jog.2023.73}, number={278}, journal={Journal of Glaciology}, author={Jouvet, Guillaume and Cordonnier, Guillaume}, year={2023}, pages={1941–1955}}
%




\end{document}